\documentclass[10pt,twocolumn,letterpaper]{article}

\usepackage{cvpr}
\usepackage{times}
\usepackage{epsfig}
\usepackage{graphicx}
\usepackage{amsmath}
\usepackage{amssymb}

\usepackage{pdflscape}
\usepackage{multirow}
\usepackage[table,xcdraw]{xcolor}

\usepackage{flushend} % 参考文献双栏平均
\usepackage[numbers, sort]{natbib}

\usepackage[symbol]{footmisc}

\usepackage{perpage} %the perpage package
\MakePerPage{footnote} %the perpage package command

\usepackage[breaklinks=true,bookmarks=false]{hyperref}

\cvprfinalcopy % *** Uncomment this line for the final submission

\def\cvprPaperID{****} % *** Enter the CVPR Paper ID here

\begin{document}

%%%%%%%%% TITLE
\title{Method Towards CVPR 2021 Image Matching Challenge}

\author{Xiaopeng Bi\footnotemark[2],  Yu Chen\footnotemark[2],  Xinyang Liu\footnotemark[2],  Dehao Zhang\footnotemark[2],  Ran Yan, Zheng Chai, Haotian Zhang, Xiao Liu\footnotemark[4]\\
Megvii Inc. Research 3D\\
{\tt\small \{bixiaopeng, chenyu05, liuxinyang02, zhangdehao, yanran, chaizheng, zhanghaotian, liuxiao\}@megvii.com}

}

\maketitle

\begin{abstract}
This report summarizes Megvii-3D team’s approach towards CVPR 2021 Image Matching Challenge. It presents details about our feature, matcher and outlier rejector selection. Also, we list the dataset used to train or fine-tune our models. In the last section, we provide a table to demonstrate the strategic combination as well as the final hyperparameters of each submission, including
\\

Unlimited keypoints category:
\begin{itemize}
\setlength\itemsep{-0.5em}
\item \href{https://www.cs.ubc.ca/research/image-matching-challenge/2021/submissions/36b2eb52df710ca0-613json/}{mss\_scale\_adapt\_f\_8k} 
\item \href{https://www.cs.ubc.ca/research/image-matching-challenge/2021/submissions/056bd6b4e63518ae-613json/}{disk\_scale\_8k}
\item \href{https://www.cs.ubc.ca/research/image-matching-challenge/2021/submissions/e90436771b444793-613json/}{sp\_disk\_scale\_8k}
\item \href{https://www.cs.ubc.ca/research/image-matching-challenge/2021/submissions/16b12c7ca1e5d8fe-614json/}{mss\_scale\_8k}
\end{itemize}

Restricted keypoints category:
\begin{itemize}
\setlength\itemsep{-0.5em}
\item \href{https://www.cs.ubc.ca/research/image-matching-challenge/2021/submissions/31b7d53f506e274e-613json/}{sp\_scale\_adapt\_f\_orien}
\item \href{https://www.cs.ubc.ca/research/image-matching-challenge/2021/submissions/d7503ab255cc200d-603json/}{mss\_scale\_v2}
\item \href{https://www.cs.ubc.ca/research/image-matching-challenge/2021/submissions/8074e77fe54827ca-66_1json/}{mss\_orien}
\item \href{https://www.cs.ubc.ca/research/image-matching-challenge/2021/submissions/b71a50d4ce4c9fb9-528json/}{mss\_degensac} 
\end{itemize}

\end{abstract}

\footnotetext{
\footnotemark[2] Work conducted during their internships at Megvii 3D
}
\footnotetext{
\footnotemark[4] Corresponding author
}

\section{Pipeline}

\begin{figure}[ht!]
    \centering      
    \includegraphics[width=8.5cm]{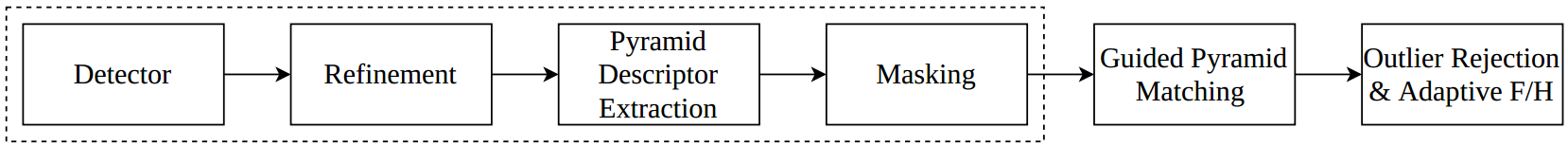}
    \caption{Method Pipeline}
    \label{fig:pipeline}
\end{figure}

\section{Feature}
\subsection{Detector \& Descriptor}
In our method, we implemented three kinds of features, pure DISK feature, pure SuperPoint\cite{superpoint_paper} feature, and a mix of SuperPoint and DISK\cite{tyszkiewicz2020disk} feature.

For SuperPoint, we additionally applied test-time homographic adaptation with 100 iterations to detect keypoints, and in some cases, we added a trained convolutional autoencoder to halve the feature dimension.

The Non-Maximum Suppression (NMS) window size and keypoints threshold was estimated theoretically based on the size of the image, to achieve a rather equally keypoints distribution across the image. Those values were then fine tuned around their initial estimates, to obtain the best performance on the validation set for each feature for each dataset.

\subsection{Pyramid Extraction}
We observed large-scale variation between images in some sets, such as the lizard in PragueParks. It troubled our matcher since the same feature under different scales may look very different. To solve this, we extracted the features in three scales and then concatenate those associated with the same keypoint. 

Other than scale, sometimes orientation is also a problem. Similar to the approach taken for the scale problem, we extracted the feature in seven orientations by affine transformations (i.e. rotate 90 degrees left and right, respectively) and perspective transformations (i.e. horizontally left/right or vertically up/down by 45 degrees, respectively), and then concatenate them.

In order to save some computational resources and make the task more efficient, we only apply scale and/or orientation in some conditions, which will be elaborated in the Appendix.

\subsection{Pre-Process \& Post-Process}
\textbf{Masking}
For the Phototoursim and GoogleUrban datasets, some dynamic objects occur frequently, which introduce unreliable and unrepeatable keypoints to our solution. To overcome this, we segmented the scene and masked out some classes, including Person, Sky, Car, Bus, and Bicycle. Except for the class Person, which was segmented by Mask-RCNN\cite{Maskrcnn} trained on COCO, all other classes were removed by the pspnet\cite{zhao2017pspnet} trained on ade20k. 

While those object segmentation nets worked well under most of the circumstances, they performed poorly when distinguishing the sculpture from humans; thus we additionally trained a binary classifier to take pedestrians apart from sculptures. 

Moreover, in order to preserve the details of buildings, when the masking was enabled, we eroded the area masked by a 5x5 kernel for sky and 3x3 for the other classes.
\\

\textbf{Refinement}
We applied an argsoftmax function for keypoints refinement with a radius of 2.
\\

% \textbf{Image Inpainting}
% TODO................

\section{Matching}
We trained the SuperGlue\cite{sarlin20superglue} together with its official feature extractor SuperPoint\cite{superpoint_paper} in an end-to-end manner on the MegaDepth dataset with IMW2021 competition testset removed. More specifically, we splited the original SuperPoint\cite{superpoint_paper} into two networks, the first one was fixed with the official weights to extract the keypoints from the image, while the other one was fine-tuned to provide descriptions. However, we found this adjustment advanced the model performance slightly, since the SuperGlue\cite{sarlin20superglue} can already match the given points well. 

Furthermore, we take the advantage of the latest feature DISK\cite{tyszkiewicz2020disk} instead of SuperPoint\cite{superpoint_paper}, which built upon a simple CNN backbone. For the SuperGlue\cite{sarlin20superglue} matcher compatible with DISK\cite{tyszkiewicz2020disk}, we only trained the matcher part, while directly used the official weights of DISK\cite{tyszkiewicz2020disk} and did not conduct any fine-tuning on it. We leveraged DISK\cite{tyszkiewicz2020disk} to extract $2,048$ points for each image in the training phase, and $8,000$ points in the testing phase. In this way, as shown in Table.\ref{table:1 superglue}, the AUC could be improved significantly. The evaluation was conducted following the methodology in the SuperGlue\cite{sarlin20superglue} paper.

We have trained an indoor and an outdoor version for the SuperGlue matcher compatible with either SuperPoint or DISK, but we did not find the indoor one help with performance advancement; thus we stuck with the outdoor weights for this competition.

\begin{table}[ht!]
\resizebox{\columnwidth}{!}{%
\begin{tabular}{|l|c|c|c|}
\hline
                                              & \multicolumn{3}{c|}{Exact AUC} \\ \hline
                                              & 5°       & 10°      & 20°      \\ \hline
SuperPoint + SuperGlue (Official)             & 38.72    & 59.13    & 75.81    \\ \hline
SuperPoint + SuperGlue (Our trained, outdoor) &  38.88    &   59.27   &  75.71  \\ \hline
DISK + SuperGlue (Outdoor)                    & 42.72    & 62.54    & 97.68    \\ \hline
\end{tabular}}
\caption{SuperGlue Weights Evaluation}
\label{table:1 superglue}
\end{table}

\textbf{Guided Pyramid Matching} Only when the number of matches found by SuperGlue was less than 100, SuperGlue matching would be applied on the pyramid extraction results, i.e. multiple scales and/or multiple orientations, we might combine the matches in different scales (\textit{ALL}) or trust the one with the most number of matches (\textit{MAX}).
\section{Outlier Rejection \& Adaptive F/H}
Based on our experiments, DegenSAC\cite{Mishkin2015MODS} outperformed other outlier rejection methods in most cases. So we applied DegenSAC to find the fundamental matrix and under certain circumstances, the homography simultaneously.

Inspired by ORB-SLAM\cite{ORBSLAM3_2020}, different transformation matrix should be selected for different scenes and then be decomposed to get pose. Although the process is fixed in the offical bankend, considering different transformation matrix still helps us to filter matching outliers. In the GoogleUrban dataset, for example, many correct matches are on the flat buildings along the street, but many wrong matches are on the ground or on the isolation belt of the road. Suffering from the weak constraint of the F matrix, some wrong matches also could pass the filter (the distance to the epipolar line is less than the threshold). While if the H matrix is selected, only the correct matches on the street-side flat building is retained. Even if the pose is decomposed by calculating the F matrix later in the offical process, the accuracy can still be improved 
since some wrong matches could be removed by adaptive F/H strategy. For the method implementation, we refer to the ORB-SLAM\cite{ORBSLAM3_2020}, respectively calculating the F matrix and the H matrix, as well as their scores (SF and SH), which are related to the symmetric transfer error less than the threshold. After obtaining SF and SH, calculate RH=SH/(SH+SF), and then we select H matrix if RH is greater than 0.45. Otherwise, we select the F matrix, which s called adaptive F-H policy. In the Pragueparks dataset, there are not many these kind of cases, so th
e performance improvement was not obvious by applying adaptive F/H. In the Phototourism dataset, most of the non-planar matches are correct matches, so applying adaptive F/H led to the correct non-planar matches being removed and thus worsen the accuracy.

\section{Conclusion}
This report provided Megvii-3D team's strategies towards CVPR2021 IMW competition for both the unlimited keypoints category and the restricted one. \\

\textbf{Limitations of our method}

From our analysis on the corner cases, we realized that a simple feature-matcher-filter solution could not suffice all conditions, without multiple scales and/or multiple orientations matching augmentation. We observed there are image pairs with scale difference more than a factor of 3, more than 45 degree of rotation, or large perspective transformation. A single level solution can not work them out by any means.

While those cases are rare but still possible in some applications for real life, for example, when you are using an AR map guider and conducting a sharp turn-around or you suddenly fail down, visual localizer is likely to fail. However, we argue that those extreme cases could be leveraged by compensations from other type of data or sensors, such as the IMU, GPS, and QR code positioning system. On the other hand, the performance of current feature-matcher might be saturated already since the convolutional neural network block has its own limitations, including the limited receptive field, hard to break the spatial relationships between pixels, cannot model shape deformations (though we have DCNs now), etc.

Despite that many excellent researchers will continue on improving the accuracy, robustness together with the generalization of feature extraction and matching, it might be doubted that stereo matching plays a dominant role in optimizing the performance of the whole visual localization system.

{\small
\bibliographystyle{IEEEtran}
\bibliography{egbib}
}

\newpage
\begin{landscape}

\section*{Appendix: Details about each Submission}

% Please add the following required packages to your document preamble:
% \usepackage{multirow}
% \usepackage[table,xcdraw]{xcolor}
% If you use beamer only pass "xcolor=table" option, i.e. \documentclass[xcolor=table]{beamer}
\begin{table}[ht!]
\centering
\resizebox{\columnwidth}{!}{%

\begin{tabular}{|c|c|c|c|c|c|c|c|c|c|c|c|c|c|c|c|c|c|c|}
\hline
                                                    &                                     &                                                                                  & \multicolumn{9}{c|}{\textbf{Keypoints}}                                                                                                                                                                                                                                                                                                                                                                                                                                                                                                           & \multicolumn{4}{c|}{\textbf{Matching (SuperGlue)}}                                                                                                                                                           & \multicolumn{3}{c|}{\textbf{RANSAC (DEGENSAC)}}                                                                            \\ \cline{4-19} 
\multirow{-2}{*}{\textbf{Submission}}               & \multirow{-2}{*}{\textbf{Dataset}}  & \multirow{-2}{*}{\textbf{\begin{tabular}[c]{@{}c@{}}Image \\ Size\end{tabular}}} & \textbf{\begin{tabular}[c]{@{}c@{}}Detector \&\\  Descriptor\end{tabular}} & \textbf{\begin{tabular}[c]{@{}c@{}}Keypoint \\ Threshold\end{tabular}} & \textbf{NMS} & \textbf{Weights}  & \textbf{\begin{tabular}[c]{@{}c@{}}Auto \\ Encoder\end{tabular}} & \textbf{Pyramid}                                                                            & \textbf{\begin{tabular}[c]{@{}c@{}}\#Keypoint \\ Nums\end{tabular}} & \textbf{\begin{tabular}[c]{@{}c@{}}\#Keypoints  \\ after Mask\end{tabular}} & \textbf{Mask}                       & \textbf{Threshold} & \textbf{Weights} & \textbf{Pyramid}                                                                            & \textbf{\begin{tabular}[c]{@{}c@{}}Sinkhorn \\ Iteration\end{tabular}} & \textbf{\begin{tabular}[c]{@{}c@{}}Threshold\\ (stereo/mv)\end{tabular}} & \textbf{\#Iterations}       & \textbf{Adapt FH} \\ \hline
{\color[HTML]{212529} }                             & {\color[HTML]{212529} Phototourism} & 1600                                                                             &                                                                            & 0.0005                                                                 & 4            & official          & ×                                                                & scale (ALL)                                                                                 & 5000                                                                & 2048                                                                        & {\color[HTML]{4D5156} $\checkmark$} & 0.2                & official         & scale (ALL)                                                                                 & 150                                                                    & 1.1/1.1                                                                  & 100k                        & ×                 \\ \cline{2-3} \cline{5-19} 
{\color[HTML]{212529} }                             & {\color[HTML]{212529} PragueParks}  & 2048                                                                             &                                                                            & 0.0005                                                                 & 4            & official          & ×                                                                & scale (ALL)                                                                                 & 5000                                                                & 2048                                                                        & ×                                   & 0.2                & official         & scale (ALL)                                                                                 & 150                                                                    & 2.5/2.5                                                                  & 100k                        & ×                 \\ \cline{2-3} \cline{5-19} 
\multirow{-3}{*}{{\color[HTML]{212529} mssscalev2}} & {\color[HTML]{212529} GoogleUrban}  & 1600                                                                             & \multirow{-3}{*}{SuperPoint}                                               & 0.0005                                                                 & 4            & official          & ×                                                                & scale (ALL)                                                                                 & 5000                                                                & 2048                                                                        & {\color[HTML]{4D5156} $\checkmark$} & 0.2                & official         & scale (ALL)                                                                                 & 150                                                                    & 1.1/1.2                                                                  & 100k                        & ×                 \\ \hline
                                                    & {\color[HTML]{212529} Phototourism} & 2000                                                                             &                                                                            & 0.0005                                                                 & 4            & official          & ×                                                                & scale (ALL)                                                                                 & 8000                                                                & {\color[HTML]{172B4D} 3586}                                                 & {\color[HTML]{4D5156} $\checkmark$} & 0.2                & official         & scale (ALL)                                                                                 & 150                                                                    & 0.9/0.9                                                                  & 1000k                       & ×                 \\ \cline{2-3} \cline{5-19} 
                                                    & {\color[HTML]{212529} PragueParks}  & 2048                                                                             &                                                                            & 0.0005                                                                 & 4            & official          & ×                                                                & scale (ALL)                                                                                 & 8000                                                                & {\color[HTML]{172B4D} 3586}                                                 & ×                                   & 0.2                & official         & scale (ALL)                                                                                 & 150                                                                    & 2.2/2.7                                                                  & 1000k                       & ×                 \\ \cline{2-3} \cline{5-19} 
\multirow{-3}{*}{mss\_scale\_adapt\_f\_8k}          & {\color[HTML]{212529} GoogleUrban}  & 1600                                                                             & \multirow{-3}{*}{SuperPoint}                                               & 0.0005                                                                 & 4            & retrain           & ×                                                                & scale (ALL)                                                                                 & 8000                                                                & {\color[HTML]{172B4D} 3586}                                                 & {\color[HTML]{4D5156} $\checkmark$} & 0.2                & retrain          & scale (ALL)                                                                                 & 150                                                                    & 0.8/0.8                                                                  & 1000k                       & only mv           \\ \hline
                                                    & {\color[HTML]{212529} Phototourism} & 1600                                                                             &                                                                            & 0.0005                                                                 & 4            & official          & ×                                                                & orien (ALL)                                                                                 & 5000                                                                & 2048                                                                        & {\color[HTML]{4D5156} $\checkmark$} & 0.2                & official         & orien (ALL)                                                                                 & 150                                                                    & 1.1/1.1                                                                  & 100k                        & ×                 \\ \cline{2-3} \cline{5-19} 
                                                    & {\color[HTML]{212529} PragueParks}  & 2048                                                                             &                                                                            & 0.0005                                                                 & 4            & official          & ×                                                                & orien (ALL)                                                                                 & 5000                                                                & 2048                                                                        & ×                                   & 0.2                & official         & orien (ALL)                                                                                 & 150                                                                    & 2.5/2.5                                                                  & 100k                        & ×                 \\ \cline{2-3} \cline{5-19} 
\multirow{-3}{*}{mss\_orien}                        & {\color[HTML]{212529} GoogleUrban}  & 1600                                                                             & \multirow{-3}{*}{SuperPoint}                                               & 0.0005                                                                 & 4            & official          & ×                                                                & orien (ALL)                                                                                 & 5000                                                                & 2048                                                                        & {\color[HTML]{4D5156} $\checkmark$} & 0.2                & official         & orien (ALL)                                                                                 & 150                                                                    & 1.1/1.2                                                                  & 100k                        & ×                 \\ \hline
                                                    & {\color[HTML]{212529} Phototourism} & 1600                                                                             &                                                                            & 0.0005                                                                 & 4            & official          & {\color[HTML]{4D5156} $\checkmark$}                              & ×                                                                                           & 5000                                                                & 2048                                                                        & {\color[HTML]{4D5156} $\checkmark$} & 0.2                & official         & ×                                                                                           & 150                                                                    & 1.1/1.1                                                                  & 100k                        & ×                 \\ \cline{2-3} \cline{5-19} 
                                                    & {\color[HTML]{212529} PragueParks}  & 2048                                                                             &                                                                            & 0.0005                                                                 & 4            & official          & {\color[HTML]{4D5156} $\checkmark$}                              & ×                                                                                           & 5000                                                                & 2048                                                                        & ×                                   & 0.2                & official         & ×                                                                                           & 150                                                                    & 2.5/2.5                                                                  & 100k                        & ×                 \\ \cline{2-3} \cline{5-19} 
\multirow{-3}{*}{mss\_degensac}                     & {\color[HTML]{212529} GoogleUrban}  & 1600                                                                             & \multirow{-3}{*}{SuperPoint}                                               & 0.0005                                                                 & 4            & official          & {\color[HTML]{4D5156} $\checkmark$}                              & ×                                                                                           & 5000                                                                & 2048                                                                        & {\color[HTML]{4D5156} $\checkmark$} & 0.2                & official         & ×                                                                                           & 150                                                                    & 1.2/1.2                                                                  & 100k                        & ×                 \\ \hline
                                                    & {\color[HTML]{212529} Phototourism} & 1600                                                                             &                                                                            & ×                                                                      & 3            & official          & ×                                                                & scale (ALL)                                                                                 & 10000                                                               & 8000                                                                        & {\color[HTML]{4D5156} $\checkmark$} & 0.7                & retrain          & scale (ALL)                                                                                 & 100                                                                    & 1.1/1.1                                                                  & {\color[HTML]{172B4D} 500k} & ×                 \\ \cline{2-3} \cline{5-19} 
                                                    & {\color[HTML]{212529} PragueParks}  & 1600                                                                             &                                                                            & ×                                                                      & 7            & official          & ×                                                                & scale (ALL)                                                                                 & 10000                                                               & 8000                                                                        & ×                                   & 0.7                & retrain          & scale (ALL)                                                                                 & 100                                                                    & 1.5/2.5                                                                  & {\color[HTML]{172B4D} 500k} & ×                 \\ \cline{2-3} \cline{5-19} 
\multirow{-3}{*}{disk\_scale\_8k}                   & {\color[HTML]{212529} GoogleUrban}  & 1600                                                                             & \multirow{-3}{*}{disk}                                                     & ×                                                                      & 4            & official          & ×                                                                & scale (ALL)                                                                                 & 10000                                                               & 8000                                                                        & {\color[HTML]{4D5156} $\checkmark$} & 0.7                & retrain          & scale (ALL)                                                                                 & 100                                                                    & 1.1/1.5                                                                  & {\color[HTML]{172B4D} 500k} & ×                 \\ \hline
                                                    & {\color[HTML]{212529} Phototourism} & 1600                                                                             &                                                                            & 0.0005                                                                 & 4            & retrain(only mv)  & ×                                                                & \begin{tabular}[c]{@{}c@{}}orien (ALL),\\ only stereo\end{tabular}                          & 5000                                                                & 2048                                                                        & {\color[HTML]{4D5156} $\checkmark$} & 0.2                & retrain(only mv) & \begin{tabular}[c]{@{}c@{}}orien (ALL),\\ only stereo\end{tabular}                          & 150                                                                    & 1.1/1.1                                                                  & 100k                        & ×                 \\ \cline{2-3} \cline{5-19} 
                                                    & {\color[HTML]{212529} PragueParks}  & 2048                                                                             &                                                                            & 0.0005                                                                 & 4            & official          & ×                                                                & \begin{tabular}[c]{@{}c@{}}orien (ALL),\\ only stereo\\ scale (ALL),\\ only mv\end{tabular} & 5000                                                                & 2048                                                                        & ×                                   & 0.2                & official         & \begin{tabular}[c]{@{}c@{}}orien (ALL),\\ only stereo\\ scale (ALL),\\ only mv\end{tabular} & 150                                                                    & 2.5/2.5                                                                  & 100k                        & ×                 \\ \cline{2-3} \cline{5-19} 
\multirow{-3}{*}{sp\_scale\_adapt\_f\_orien}        & {\color[HTML]{212529} GoogleUrban}  & 1600                                                                             & \multirow{-3}{*}{SuperPoint}                                               & 0.0005                                                                 & 4            & official          & ×                                                                & \begin{tabular}[c]{@{}c@{}}scale (ALL),\\ only mv\end{tabular}                              & 5000                                                                & 2048                                                                        & {\color[HTML]{4D5156} $\checkmark$} & 0.2                & official         & \begin{tabular}[c]{@{}c@{}}scale (ALL),\\ only mv\end{tabular}                              & 150                                                                    & 1.1/1.2                                                                  & 100k                        & only mv           \\ \hline
                                                    & {\color[HTML]{212529} Phototourism} & 2000                                                                             &                                                                            & 0.0005                                                                 & 4            & official          & ×                                                                & scale (ALL)                                                                                 & 8000                                                                & {\color[HTML]{172B4D} 3586}                                                 & {\color[HTML]{4D5156} $\checkmark$} & 0.2                & official         & scale (ALL)                                                                                 & 150                                                                    & 0.9/0.9                                                                  & 1000k                       & ×                 \\ \cline{2-3} \cline{5-19} 
                                                    & {\color[HTML]{212529} PragueParks}  & 2048                                                                             &                                                                            & 0.0005                                                                 & 4            & official          & ×                                                                & scale (ALL)                                                                                 & 8000                                                                & {\color[HTML]{172B4D} 3586}                                                 & ×                                   & 0.2                & official         & scale (ALL)                                                                                 & 150                                                                    & 2.2/2.7                                                                  & 1000k                       & ×                 \\ \cline{2-3} \cline{5-19} 
\multirow{-3}{*}{mss\_scale\_8k}                    & {\color[HTML]{212529} GoogleUrban}  & 1600                                                                             & \multirow{-3}{*}{SuperPoint}                                               & 0.0005                                                                 & 4            & retrain           & ×                                                                & scale (ALL)                                                                                 & 8000                                                                & {\color[HTML]{172B4D} 3586}                                                 & {\color[HTML]{4D5156} $\checkmark$} & 0.2                & retrain          & scale (ALL)                                                                                 & 150                                                                    & 0.8/0.8                                                                  & 1000k                       & ×                 \\ \hline
                                                    & {\color[HTML]{212529} Phototourism} & 1600/1600                                                                        &                                                                            & 0.0005/x                                                               & 4/3          & official/official & ×                                                                & scale(ALL)/scale(ALL)                                                                       & 5000/8000                                                           & 2048/5900                                                                   & {\color[HTML]{4D5156} $\checkmark$} & 0.2/0.7            & official/retrain & scale (ALL)                                                                                 & 150/100                                                                & 1.1/1.1                                                                  & 500k                        & ×                 \\ \cline{2-3} \cline{5-19} 
                                                    & {\color[HTML]{212529} PragueParks}  & 2048/1600                                                                        &                                                                            & 0.0005/x                                                               & 4/8          & official/official & ×                                                                & scale(ALL)/scale(ALL)                                                                       & 5000/8000                                                           & 2048/5900                                                                   & ×                                   & 0.2/0.7            & official/retrain & scale (ALL)                                                                                 & 150/100                                                                & 2.5/2.5                                                                  & 500k                        & ×                 \\ \cline{2-3} \cline{5-19} 
\multirow{-3}{*}{sp\_disk\_scale\_8k}               & {\color[HTML]{212529} GoogleUrban}  & 1600/1600                                                                        & \multirow{-3}{*}{SuperPoint/disk}                                          & 0.0005/x                                                               & 4/4          & official/official & ×                                                                & scale(ALL)/scale(ALL)                                                                       & 5000/8000                                                           & 2048/5900                                                                   & {\color[HTML]{4D5156} $\checkmark$} & 0.2/0.7            & official/rerain  & scale (ALL)                                                                                 & 150/100                                                                & 1.1/1.5                                                                  & 500k                        & ×                 \\ \hline
\end{tabular}%
}

\caption{Submission Details, 
\textit{the column \textbf{pyramid}/\textbf{guided pyramid} refers to the pyramid feature extraction/guided pyramid matching. scale (ALL) denotes that we combined and used all of the matches found in all scales, orien (ALL) denotes that we combined and used all of the matches found in all orientations. }
}

\end{table}

\end{landscape}

\end{document}